\documentclass[sigconf]{acmart}

\AtBeginDocument{%
  }

\copyrightyear{2026}
\acmYear{2026}
\setcopyright{cc}
\setcctype{by}
\acmConference[ICPC '26]{34th IEEE/ACM International Conference on Program Comprehension}{April 12--13, 2026}{Rio de Janeiro, Brazil}
\acmBooktitle{34th IEEE/ACM International Conference on Program Comprehension (ICPC '26), April 12--13, 2026, Rio de Janeiro, Brazil}
\acmPrice{}
\acmDOI{10.1145/3794763.3794808}
\acmISBN{979-8-4007-2482-4/2026/04}

\usepackage{amsfonts}
\usepackage{algorithm}
\usepackage{algorithmic}
\usepackage{enumitem}
\usepackage{multirow}
\usepackage{colortbl}
\usepackage{makecell} 

\title{SCOPE: Tree-based Self-Correcting Online Log Parsing via Syntactic-Semantic Collaboration}

\author{Dongyi Fan}
\orcid{0009-0008-1345-6644}
\affiliation{%
  \institution{Zhejiang Sci-Tech University}
  \city{Hangzhou}
  \state{Zhejiang}
  \country{China}}
\email{2023110602001@mails.zstu.edu.cn}

\author{Suqiong Zhang}
\orcid{0000-0002-6365-0727}
\affiliation{%
  \institution{Zhejiang Sci-Tech University}
  \city{Hangzhou}
  \state{Zhejiang}
  \country{China}}
\email{zhangsuqiong22@gmail.com}

\author{Lili He}
\authornote{Corresponding author}
\affiliation{%
  \institution{Zhejiang Sci-Tech University}
  \city{Hangzhou}
  \state{Zhejiang}
  \country{China}}
\email{llhe@zju.edu.cn}

\author{Ming Liu}
\affiliation{%
  \institution{Zhejiang Sci-Tech University}
  \city{Hangzhou}
  \state{Zhejiang}
  \country{China}}
\email{2024110602003@mails.zstu.edu.cn}

\author{Yifan Huo}
\affiliation{%
  \institution{Zhejiang Sci-Tech University}
  \city{Hangzhou}
  \state{Zhejiang}
  \country{China}}
\email{2024010602002@mails.zstu.edu.cn}

\begin{document}
\begin{abstract}
Log parsing is a critical step for automated log analysis in complex systems. Traditional heuristic-based methods offer high efficiency but are limited in accuracy due to overlooking semantic context. In contrast, recent LLM-based parsers improve accuracy via semantic understanding but incur high latency from frequent model calls. To address this, we propose SCOPE, the first self-correcting online log parsing method that integrates the strengths of both heuristic and LLM-based paradigms. SCOPE introduces a novel bi-directional tree structure that enables efficient template matching from both forward and reverse directions, resulting in a higher overall matching rate. Additionally, it adopts a two-stage syntactic-semantic collaboration framework: a lightweight NLP model first utilizes part-of-speech (POS) information for syntax-based matching, while the LLM is selectively invoked as a fallback to handle semantically complex cases when uncertainty remains. This design significantly reduces LLM API usage while maintaining high accuracy, achieving a balance between efficiency and effectiveness. Extensive evaluations on diverse benchmark datasets show that SCOPE outperforms state-of-the-art methods in both accuracy and efficiency. The implementation and datasets are publicly released to facilitate further research.
\end{abstract}



\maketitle

\section{Introduction}

Log messages are textual records of events, transactions, or activities generated by software systems at runtime~\cite{logpractice,beck2025logparsing,skopik2021loganalysis,Zhou2018}, providing crucial insights into system behavior and state. Due to their high human readability, logs serve as a primary data source for a variety of downstream tasks, including anomaly detection~\cite{loganomalynoparse, deeptralog,he2016experience}, root cause analysis~\cite{du2017deeplog,lee2023eadro}, and system diagnostics~\cite{le2022log}.
However, with the increasing scale and complexity of modern systems, the volume and diversity of log data have grown substantially, making manual analysis infeasible. To address this, automated log analysis pipelines have been developed, with \textit{log parsing} as a foundational step \cite{he2016evaluation}. An accurate log parser is always in high demand because it provides reliable input to ensure effective downstream analysis. Specifically, log parsing is the task of converting a raw log message into a specific log template associated with the corresponding parameters via extracting constant and variable parts~\cite{autologana}. For example, the raw log message \texttt{"Session started on port 62267"} can be parsed into the template \texttt{"Session started on port <*>"}, here the wildcard symbol \verb|<*>| is used to represent the variable port number.


Traditional log parsing methods have evolved from handcrafted regular expressions to structural pattern mining, and more recently to approaches that incorporate syntactic and semantic analysis for improved generalization and robustness. Regular expression-based methods \cite{xu2009detecting} are increasingly outdated due to their high maintenance cost and poor adaptability to new log formats. Structure-based mining approaches \cite{drain,spell,Logram,nagappan2010abstracting} extract features such as token count, frequency, and position to identify constant tokens, and demonstrate strong efficiency in practical deployments. Syntax-based methods \cite{posparser,pokharel2023hybrid} utilize syntactic information, including part-of-speech (POS) tags and parsing trees, to better distinguish constants from variables. Semantic-based approaches further improve accuracy by leveraging pretrained models such as RoBERTa \cite{logPPT} and large language models (LLMs) \cite{logbatcher, Lunar, LILAC, divlog, Logparser-LLM} to extract templates from a semantic perspective.


Despite significant advancements in log parsing methodologies, existing approaches still face several challenges. On one hand, heuristic log structure-based methods heavily rely on handcrafted rules and domain-specific knowledge. As a result, their effectiveness tends to degrade when applied to diverse log types, often requiring manual hyperparameter tuning for each dataset, which limits their generalizability and practical applicability. On the other hand, semantic-based methods that leverage machine learning models typically demand substantial training efforts, including training from scratch or fine-tuning pre-trained language models on labeled data—which is often scarce and expensive to obtain. While LLMs eliminate the need for task-specific training and enhance generalization, frequent invocations lead to considerable latency and financial overhead. 

To address the aforementioned limitations, we propose SCOPE, a tree-based \underline{S}elf-\underline{C}orrecting \underline{O}nline log \underline{P}arsing with syntactic-s\underline{E}mantic collaboration framework. Specifically, SCOPE employs a bi-directional parsing tree (BDPT) to match log messages in both forward and reverse directions. When a message cannot be matched locally, a global template search is activated through the Pool-based Template Matching Process (PTMP). During both BDPT and PTMP matching stages, a two-stage syntactic–semantic collaborative module, termed the NLP-based Extractor (NLPE), is employed for template extraction. In this module, a lightweight NLP model leverages POS information to handle most template matching tasks, while the LLM is selectively invoked only when syntactic cues are insufficient. This hierarchical and adaptive design allows SCOPE to achieve high accuracy and efficiency while maintaining strong adaptability across diverse log formats.

We conducted a comprehensive evaluation on 14 large-scale datasets from Loghub-2.0~\cite{loghub}. The results demonstrate that SCOPE achieves the highest average accuracy across all performance metrics, outperforming state-of-the-art heuristic parsers such as Drain and Brain by an average of 16.1\% and 49.5\% in GA and PA, and by 94.8\% and 167.7\% in the F1 scores of grouping and template accuracy, respectively. Compared with the best-performing LLM-based parsers, SCOPE achieves average improvements of 3.8\%, 13.5\%, 1.5\%, and 12.0\% in GA, PA, FGA, and FTA, respectively. In terms of efficiency, SCOPE demonstrates remarkable parsing speed, surpassing the most efficient baseline, Drain. Moreover, it requires only about one-sixth as many LLM calls as other LLM-based baselines, significantly reducing the overhead associated with LLM invocations.

The main contributions of this paper are as follows:
\begin{itemize}
    \item We propose SCOPE, the first unified log parsing framework that integrates traditional heuristic techniques with LLM-based semantic analysis. By leveraging the efficiency of heuristic methods and the accuracy of LLMs, SCOPE achieves state-of-the-art results in both aspects.
    
    \item We propose the first parsing method using a Bi-Directional Parse Tree (BDPT) with a dynamic depth design. By performing template matching in both forward and reverse directions, the approach enhances the ability to capture templates. Additionally, the self-correcting mechanism, which replaces previously incorrect branches with new ones, further improves the overall matching accuracy.
    
    \item We introduce a novel two-stage syntax–semantic matching framework that combines syntax-based POS tagging with semantic-based LLM usage, thereby significantly improving overall efficiency by reducing both the number and cost of LLM invocations.
    
    \item We release our code to the public to promote transparency, reproducibility, and future research.
\end{itemize}

\section{Motivation}

\textbf{Lack of integration between traditional parsing and LLM-based methods.}
Traditional log parsers, particularly those based on structural heuristics, provide high efficiency and are well-suited for online processing~\cite{logsurvey}. In contrast, LLM-based parsers exhibit strong semantic understanding, significantly enhancing the effectiveness of log parsing~\cite{llmlogsurvey}. Nevertheless, existing research has yet to effectively integrate the strengths of both paradigms into a unified framework. For example, although the LLM-based parser LILAC~\cite{LILAC} employs a tree structure for preliminary template matching, the tree serves merely as a cache, with all template extraction offloaded to the LLM, thereby limiting its adaptability and self-evolution capabilities.

\textbf{Over-reliance on LLMs introduces inefficiencies.}
Recent LLM-based log parsing methods have achieved state-of-the-art accuracy by leveraging the semantic capabilities of models such as GPT~\cite{logeval,Logparser-LLM,llmlogsurvey}. Most approaches first cluster similar log messages and then apply LLMs to extract templates for each cluster. However, this strategy often results in redundant or unnecessary LLM invocations, particularly for simple or unambiguous log messages (e.g., those containing only a single token). Consequently, latency and computational cost become significant obstacles for practical deployment.

\textbf{Insufficient utilization of lightweight syntactic cues.}
Log parsers such as XDrain~\cite{liu2024xdrain} and PosParse~\cite{posparser} demonstrate that syntactic features, including part-of-speech (POS) tags, are highly effective for identifying log structure and distinguishing between variable and constant tokens. Since these features are lightweight and require neither model training nor inference, they are particularly well-suited for performance-sensitive systems. Nevertheless, current LLM-driven approaches often overlook these syntactic cues in early-stage processing, highlighting an opportunity to integrate them for improved efficiency.

\section{Methodology}
\subsection{Overview}
\begin{figure*}[htbp]
    \centering
    \includegraphics[width=\linewidth]{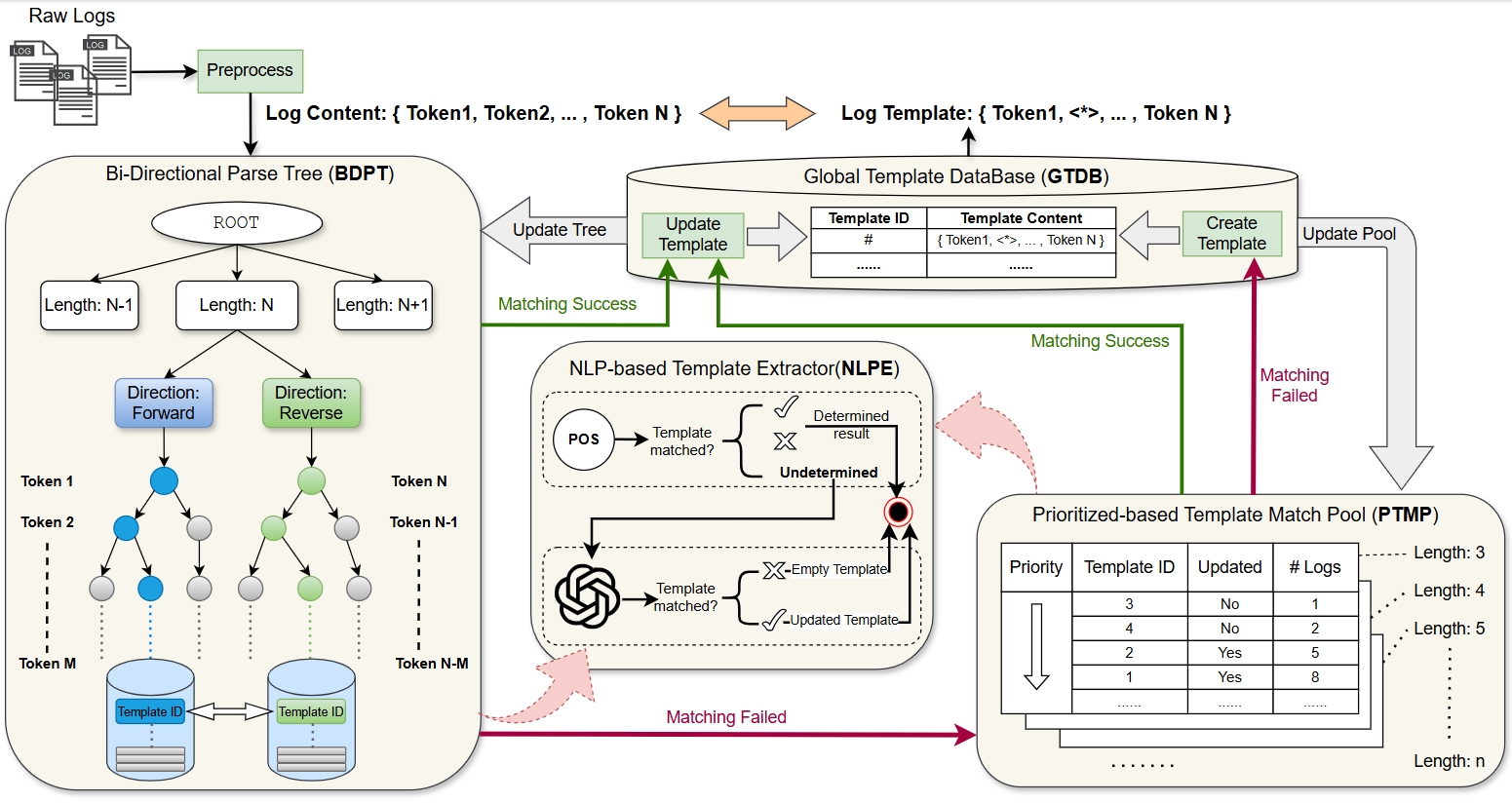}
    \caption{Architecture of SCOPE}
    \label{fig:arch}
\end{figure*}

The SCOPE framework primarily consists of four components: a preprocessing module, a Bi-Directional Parse Tree (BDPT), a Priority-based Template Match Pool (PTMP), and an NLP-based Template Extractor (NLPE), which work collaboratively to enable efficient log parsing. The overall architecture of SCOPE is illustrated in Figure~\ref{fig:arch}.
When a log message arrives and is processed by the preprocessing module, it is first sent to BDPT, which attempts template matching by traversing the parse tree in both forward and reverse directions using node-by-node comparison strategy.
    If no matching template is found in either direction of the tree, this typically indicates the emergence of a new template or an error in previously extracted ones. In such cases, the PTMP is invoked, employing a similarity-based approach to search across all known templates. A successful match leads to the correction of previously misidentified templates and their update in the parse tree. If no suitable template is found, the message is treated as a new template and added to the parse tree. During processing in both BDPT and PTMP, the NLPE is utilized to effectively distinguish between constant and variable parts of log messages, thereby enhancing the accuracy of template extraction.

\subsection{Preprocessing}

From relevant literature and practical experience, log preprocessing can improve template extraction accuracy \cite{drain}. Therefore, before applying advanced analysis methods, raw log messages are preprocessed upon arrival, which primarily includes token splitting and common variable filtering. Whitespace is commonly used for token splitting, SCOPE further treats punctuation symbols(such as "=", ":", and ",") as individual tokens. This design offers two benefits: (1) it increases the diversity of token sequences, which enhances candidate template indexing based on log length; and (2) punctuation tokens often appear as constants in log messages, making them valuable cues for the subsequent syntax based template matching process. One example of token splitting is as follows: the log entry \texttt{"authentication failure; user=guest"} is split into six tokens stored in a list: \texttt{["authentication", "failure", ";", "user", "=", "guest"]}. Common variable filtering is performed using regular expressions to replace dynamic variables with placeholders. Specifically, regular expressions are utilized to identify and replace common variable patterns such as email address, numeric values, and system identifiers like process IDs or user IDs. The regular expressions employed in this step are often straightforward, as they are designed to match individual tokens rather than entire log messages. Additionally, unlike other methods that require designing separate regular expressions for each dataset, SCOPE defines a common set of expressions that can be applied across various datasets, such as those for IP addresses, file paths, and timestamps.

\subsection{BDPT: Bi-Directional Parse Tree}
Based on our observations, the number of variable tokens in log messages is significantly smaller than that of constant tokens. Moreover, these variable tokens are typically concentrated either in the first half or the second half of the message. This insight motivated our design of BDPT: a bi-directional parse tree that captures template patterns from both the forward and reverse directions. This approach enables more accurate matching compared to a single-directional tree structure, which assumes that only constant tokens appear in the first part of the message, thereby improving the template matching success rate.
Building upon the Drain method~\cite{drain}, BDPT enhances log parsing by introducing a bi-directional parse tree with dynamic depth. The tree is structured into five layers. The top layer contains a single root node, serving as the entry point. The second layer includes nodes representing the token length $N$ of incoming log messages. Each of these nodes connects to the third layer, which branches into two directions: forward and reverse, indicating the direction of template traversal. The fourth layer stores log templates as branches constructed in both directions—one from the beginning to the end and the other in reverse—while sharing the same underlying template. The depth $M$ of each branch is dynamically determined based on the template length $N$. Finally, the fifth layer consists of log template groups, where templates with the same length $N$ and identical sequences of the first $M$ matched tokens are organized under the same group.

\subsubsection {Parse Tree Construct}
The root node is initialized automatically. When receiving a log message, its token length $N$ has been determined during preprocessing, a corresponding length node is created if it does not exist. Under this node, two directional nodes, forward and reverse, are maintained. For a new log message with no matching template, a shared template of length $N$ is created, along with two branches constructed from the first and last $M$ tokens, respectively. These branches are placed under the corresponding directional nodes, and the template is stored in the associated template groups. The dynamic tree branch depth $M$ is calculated as follows:
\begin{equation}
M = 
\begin{cases}
\frac{N + 1}{2}, & \text{if } N \text{ is odd}, \\
\frac{N}{2} + 1, & \text{if } N \text{ is even}.
\end{cases}
\label{eq:M-definition}
\end{equation}

\begin{figure}[htbp]
\centerline{\includegraphics[width=\linewidth]{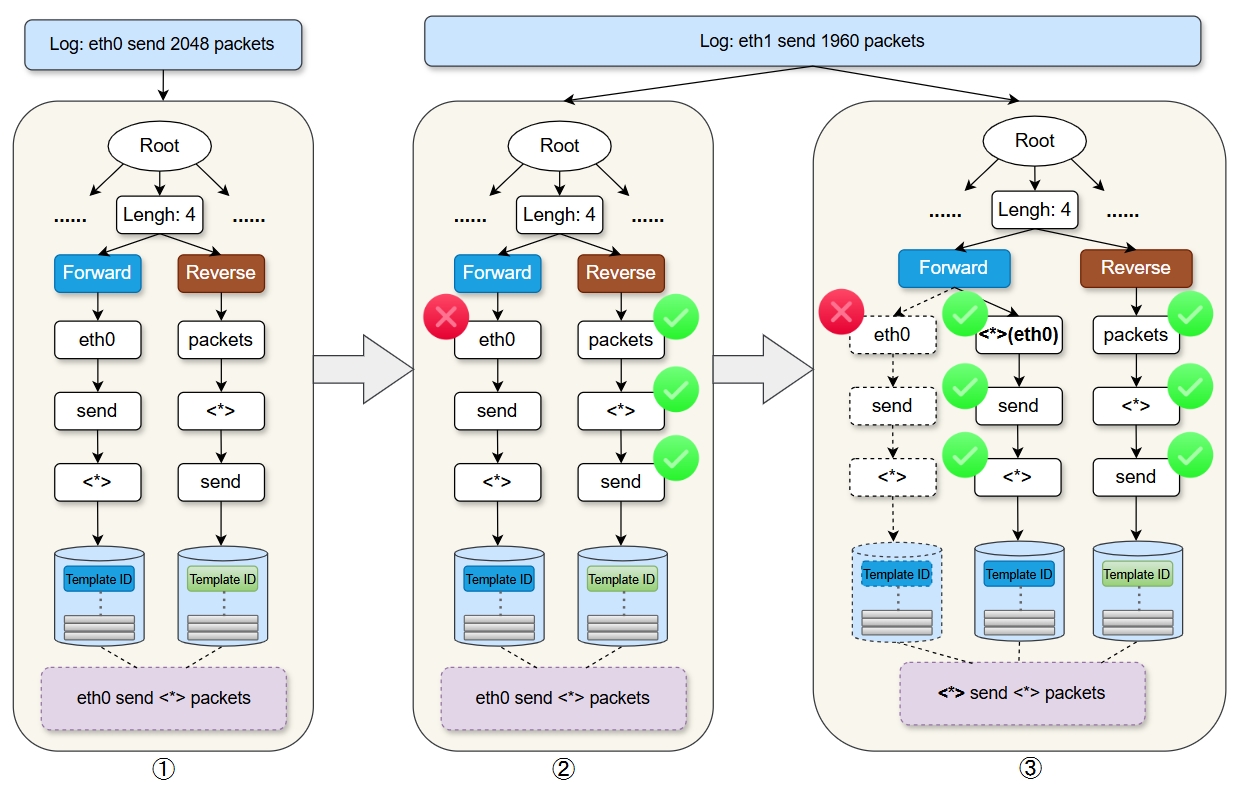}}
\caption{Example of log parsing in BDPT. 
\textcircled{1} A log template is constructed in the bi-directional parse tree. 
\textcircled{2} A new log is matched to the template in the reverse branch but is missing in the forward branch of the parse tree. 
\textcircled{3} The log template is updated and the incorrect branch is obsoleted.}

\label{fig:bdpt}
\end{figure}

With this design, at least one token is shared between the forward and reverse branches which enhances their dependency and enabling more accurate template matching through joint consideration. At this stage, no further variable extraction is performed beyond preprocessing, and the full token sequence forms the initial template. As shown in Part 1 of Figure~\ref{fig:bdpt}, the log message \texttt{"eth0 send 2048 packages"} is parsed as a new template. The numeric token \texttt{"2048"} is identified as a variable via regular expression and replaced with \verb|<*>|, resulting in a forward branch \texttt{"eth0 send <*>"} and a reverse branch \texttt{"packages <*> eth0"}. This template is then stored in the template groups of both branches.

\subsubsection {Parse Tree Match}
Template matching determines whether an existing template corresponds to an incoming log message by traversing both forward and reverse branches of the tree. As shown in Part 2 of Figure~\ref{fig:bdpt}, the log message \verb|"eth1 send 1960 packages"| is tokenized and preprocessed into \verb|"eth1 send <*> packages"|. When passed to the BDPT, the forward branch fails to match due to the difference between \verb|"eth1"| and \verb|"eth0"|. However, the reverse sequence \verb|"packages <*> send eth1"| successfully matches a reverse branch, indicating a potential match in its associated template group. A token-based similarity comparison is then performed against the templates in the group. Specifically, the similarity score is calculated as the ratio of matching tokens to total tokens and compared against a threshold of 0.5. The template with a similarity score exceeding the threshold and having the highest ratio is selected as the candidate in this branch. If both branches yield candidate templates, the final template is chosen based on the highest similarity score among all matched templates. In cases where multiple templates share the same highest similarity score, the template with fewer wildcards (<*>) is selected.

\subsubsection {Parse Tree Update}
Once a template is matched via the parse tree, a new template is generated by merging it with the incoming log message and differing tokens are replaced with \verb|<*>|. As shown in Part 3 of Figure~\ref{fig:bdpt}, this updated template replaces the previous one in the corresponding tree branches’ log groups.

\begin{center}
\setlength{\fboxsep}{4pt}
\fbox{%
  \parbox{0.95\columnwidth}{%
    \centering
    \small
    \begin{tabular}{r c c c c}
    \text{New Log Message}& : \texttt{\textbf{eth1} send <*> packages} \\
    \text{Matched Template}& : \texttt{\textbf{eth0} send <*> packages} \\
    \text{Updated Template}& : \texttt{\textbf{ <*>} send <*> packages}
    \end{tabular}
  }
}
\end{center}
When updating an existing template, the new template must be propagated to both forward and reverse branches to fix any incorrect paths. Since each parent node allows only one wildcard (\texttt{<*>}) child, conflicts arise when multiple child nodes need to be converted into wildcards. It leads to branch merging and increased system complexity. To address this, BDPT adds a new wildcard node instead of replacing the original variable token directly. Tokens following the wildcard form a new branch under it, preventing unintended merges. The original branch with the incorrect token is marked obsolete, as it will no longer match due to the wildcard’s precedence during parsing, as shown in Part 3 of Figure~\ref{fig:bdpt}. This approach avoids frequent tree modifications which reduces complexity at a minor memory cost.

\subsection{PTMP: Priority-based Template Match Pool}
The bi-directional parse tree (BDPT) improves template matching but it still fails when variable tokens appear simultaneously in both halves of a log message. To address this, the Priority-based Template Match Pool (PTMP) performs an additional global matching round over the complete template pool.
To enhance efficiency, PTMP employs a priority-based matching strategy. Within a template group sharing the same token length, priority depends on whether a template has been previously matched and updated, and on the number of log messages associated with it. Templates already matched and updated are considered stable and assigned lower priority, while those never matched or updated are deemed less reliable and given higher priority. Among templates with the same update status, those with fewer matched logs are treated as more uncertain and prioritized higher.
Formally, we define a priority tuple for each template:

\begin{equation}
p_i = (u_i,\ n_i)
\end{equation}

where $u_i \in \{0, 1\}$ indicates whether template $T_i$ has been updated, and $n_i$ is the number of log messages matched by $T_i$. The templates are sorted in ascending order of $p_i$, so that those with lower update frequency and smaller matching size appear earlier in the list. This prioritization ensures that less-certain templates are checked first during the global matching stage. In practice, one may choose to compare only the top few highest-priority candidate templates, or limit the comparison to those templates that have not been updated, thereby improving matching efficiency. The detailed procedure is given in algorithm ~\ref{alg:template-sorting}.

\begin{algorithm}[h]
\caption{Priority-Based Template Sorting}
\label{alg:template-sorting}
\begin{algorithmic}[1]
\STATE \textbf{Input:} List of templates $\mathcal{T} = \{T_1, T_2, \ldots, T_N\}$
\STATE \textbf{Output:} Sorted list $\mathcal{T'}$ with higher-priority templates at the front
\FORALL{$T_i \in \mathcal{T}$}
    \STATE $u_i \gets$ whether $T_i$ has been updated(1 if updated, 0 otherwise)
    \STATE $n_i \gets$ number of log messages matched by $T_i$
    \STATE Assign priority tuple: $p_i \gets (u_i,\ n_i)$
\ENDFOR
\STATE Create list of pairs: $\mathcal{L} \gets \{(T_i, p_i)\}_{i=1}^N$
\STATE Sort $\mathcal{L}$ by ascending $p_i$
\STATE Extract sorted templates $\mathcal{T'}$ from sorted $\mathcal{L}$
\RETURN $\mathcal{T'}$
\end{algorithmic}
\end{algorithm}

When a template is matched from the priority-sorted pool, a new template is created by merging it with the incoming log message. This updated template replaces the original one and is updated in the bi-directional parse tree. Its priority is then recalculated, and templates in the same group are re-sorted accordingly.

\subsection{NLPE: NLP-based Template Extractor}
NLPE is encapsulated as a plugin that leverages natural language processing techniques to enhance the recognition of constants and variables within log messages to help improving both accuracy and efficiency of template matching. It can be invoked by both BDPT and PTMP modules during the matching process between log messages and existing templates showed in Figure~\ref{fig:arch}.

Unlike traditional coarse-grained template matching approaches, which compute the similarity between a log message and an existing template by measuring the proportion of identical tokens at corresponding positions, and then treat them as belonging to the same template if the similarity exceeds a threshold, this naive strategy overlooks the possibility that differing tokens at the same position may actually be constants and they don't share same template even if they appear highly similar. This can lead to incorrect template matches and consequently cause template pollution. In contrast, by leveraging syntactic and semantic analysis, the NLP-based extractor (NLPE) can promptly determine whether differing tokens at the same position are fixed constants, thereby avoiding false matches. Specifically, NLPE is designed as a two-stage framework leveraging syntactic-semantic model collaboration to harness complementary strengths. In the first stage, a lightweight NLP model is employed to efficiently identify fixed constants from a syntactic perspective using part-of-speech (POS) tagging. When the NLP model cannot decisively determine whether the tokens belong to the same template, the second stage invokes a more powerful LLM to perform semantic analysis based on its superior comprehension capabilities. This synergistic collaboration enables NLPE to balance parsing accuracy and computational efficiency in online log template matching.

\subsubsection {Stage I: Syntactic Analysis with NLP Model}
An event typically consists of an event mention, event trigger words, event arguments, and their corresponding roles~\cite{ACEsurvey}. Event triggers are words, usually verbs, that indicate the occurrence of an event. Log events can be regarded as a specific type of natural language event that describes changes in system states and signals occurrences of interest to developers. Therefore, verbs in log events usually appear as constant tokens. Furthermore, according to common log-writing conventions~\cite{variableAwareLog}, punctuation marks, conjunctions (e.g., and, or), adpositions (e.g., at, to), and determiners (e.g., the, this) serve as structural components that maintain sentence formation and are also typically treated as constant tokens.

Based on these linguistic observations, the first stage of NLPE employs a lightweight NLP model (e.g., spaCy or NLTK) to perform part-of-speech (POS) tagging on log message tokens and determine whether two logs belong to different templates by examining whether differing tokens at the same position are fixed constants. Specifically, when comparing a log message with an existing template, the NLP model first assigns POS tags to each token in the log message, while the template tokens retain POS tag information from prior comparisons. During the subsequent token-by-token comparison, if the tokens at the same position differ and either token is identified as a fixed constant based on its POS tag, the comparison terminates, and the log message is classified as belonging to a new template.




In the example below, the new log message and the existing template exhibit high overall similarity; however, the tokens “send” and “received” differ at the same position. According to their POS tags, both are verbs and thus regarded as fixed constants. Consequently, the mismatch at this constant position indicates that the new log message does not belong to the existing template.


\begin{center}
\setlength{\fboxsep}{4pt}
\fbox{%
  \parbox{0.95\columnwidth}{%
    \centering
    \small
    \begin{tabular}{r c c c c}
    \text{New Log Message:} & 
    \texttt{eth0} & \texttt{\textbf{send}} & \texttt{<*>} & \texttt{packages} \\
     & \texttt{[PROPN]} & \texttt{\textbf{[VERB]}} & \texttt{[SYM]} & \texttt{[NOUN]} \\
    [1ex]
    \text{Existing Template:} & 
    \texttt{eth0} & \texttt{\textbf{received}} & \texttt{<*>} & \texttt{packages} \\
     & \texttt{[PROPN]} & \texttt{\textbf{[VERB]}} & \texttt{[SYM]} & \texttt{[NOUN]} \\
    \end{tabular}
  }
}
\end{center}


\subsubsection {Stage II: Semantic Analysis with LLM}

With powerful semantic understanding capabilities, LLMs are employed in Stage 2 to resolve ambiguous cases left unresolved by Stage 1’s syntactic analysis, particularly when tokens at the same position differ but have the same POS tag classified as undetermined. When multiple existing templates cannot be excluded by the Stage 1 NLP model syntactic analysis, only the one with the highest similarity to the new log message is selected and passed to the LLM for further semantic matching. If the LLM determines that the new log message and the selected template share the same structure, it returns the updated template accordingly. Otherwise, it returns an empty result, indicating that the new log message does not match the given template. 
Although LLMs possess strong contextual understanding capabilities, they have not been specifically fine-tuned for log parsing tasks. As a result, they still face significant challenges when dealing with the following two types of cases.

a. Identifying template mismatch.
Consider the example below:

\begin{center}
\setlength{\fboxsep}{4pt}
\fbox{%
  \parbox{0.95\columnwidth}{%
    \centering
    \small
    \begin{tabular}{r c c c c c}
      \text{New Log Message:} & 
      \texttt{\textbf{Removable}} & \texttt{base} & \texttt{files} & \texttt{:} & \texttt{<*>} \\
      \text{Existing Template:} & 
      \texttt{\textbf{Active}} & \texttt{base} & \texttt{files} & \texttt{:} & \texttt{<*>} \\
    \end{tabular}
  }
}
\end{center}

The structure may appear similar, but "Removable" and "Active" convey distinct semantic meanings, representing different states of base files. To maintain the expressiveness of the log message, these tokens should be treated as fixed constants, and the logs should be categorized under separate templates.

b. Identifying template match.
Consider the following example:

\begin{center}
\setlength{\fboxsep}{4pt}
\fbox{%
  \parbox{0.95\columnwidth}{%
    \centering
    \small
    \begin{tabular}{r c c c c c}
    \text{New Log Message:} & 
    \texttt{Failed} & \texttt{password} & \texttt{for} & \texttt{user} & \texttt{\textbf{oracle}} \\
    \text{Existing Template:} & 
    \texttt{Failed} & \texttt{password} & \texttt{for} & \texttt{user} & \texttt{\textbf{ubuntu}} \\
    \end{tabular}
  }
}
\end{center}

Although "oracle" and "ubuntu" are typically considered proper nouns and often treated as constants, in this context they represent specific instances of the abstract concept "user" and should be treated as variables. An effective LLM-based analysis should semantically understand this contextual usage in order to correctly identify such tokens as variables and determine that these log messages belong to the same template.

To address the challenges faced by LLMs in template matching, particularly in the aforementioned cases, prompt engineering plays a critical role \cite{shao2023prompting,zhang2023instruction}. Two strategies are employed to enhance the matching accuracy of LLMs. First, instead of relying on in-context learning (ICL) with diverse examples as commonly seen in prior literature, we explicitly encode semantic matching rules in natural language. This direct specification of rules allows the LLM to better understand user intent and produce more accurate results, the specific rules are list in Table~\ref{tab:prompt_rules}.

\begin{table}[ht]
\small
\renewcommand{\arraystretch}{1.3}
\caption{Rules for token classification}
\label{tab:prompt_rules}
\centering
\begin{tabular}{|m{1.6cm}|m{6.2cm}|}
\hline
\rowcolor{gray!20}
\multicolumn{1}{|c|}{\textbf{Category}} & \multicolumn{1}{c|}{\textbf{Rules}} \\
\hline
\centering\textbf{Constant} &
\vspace{8pt}
\begin{itemize}[leftmargin=*]
  \item Domain-specific term (e.g., \texttt{IPv4}).
  \item Modifier in a compound noun representing a fixed label (e.g., \texttt{Failed}).
  \item Subject in a subject–verb–object structure.
  \item Tokens expressing opposing or discrete semantics (e.g., \texttt{boot} vs. \texttt{shutdown}).
  \item Singular/plural variants are not treated as equivalent (e.g., \texttt{user} vs. \texttt{users}). 
\end{itemize} \\
\hline
\centering\textbf{Variable} &
\vspace{8pt}
\begin{itemize}[leftmargin=*]
  \item Identifiers, IPs, timestamps, user names, and other data-like tokens.
  \item Key–value patterns: retain the key and abstract the value (e.g., \texttt{user root} \textrightarrow{} \texttt{user <*>}).
  \item key:value or key=value forms: keep key, abstract value.
\end{itemize} \\
\hline
\end{tabular}
\end{table}

\begin{figure}
    \centering
    \includegraphics[width=\linewidth]{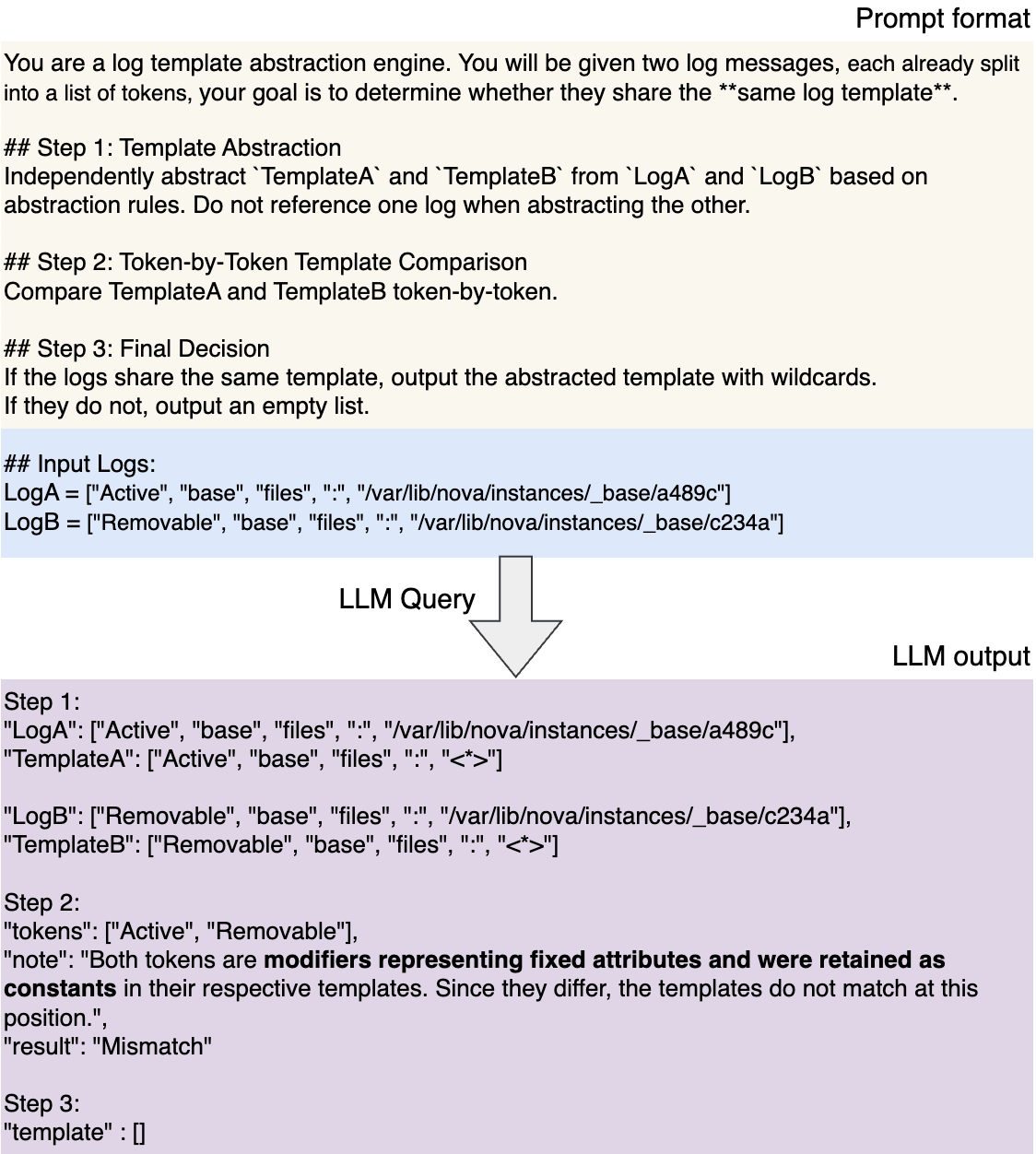}
    \caption{Prompt design illustration}
    \label{fig:prompt}
\end{figure}

Second, we adopt a chain-of-thought (CoT) prompting strategy to guide the model through a step-by-step reasoning process \cite{zhang2024lemur,wei2022chain}. Specifically, the LLM is first prompted to extract templates separately from the new log message and the existing candidate template, and then compare the two extracted templates to reach a final decision. This separate template extraction design is motivated by practical observations that LLMs tend to assume the presence of variables in log messages. When the new log message and the existing template are presented together for direct comparison, the model is more likely to mistakenly treat differing tokens as variables, leading to incorrect matches. By prompting the model to extract templates from each input independently, we encourage more objective abstraction and reduce the likelihood of such semantic bias. The prompt design is shown in Figure~\ref{fig:prompt}.

\section{Evaluation}
To evaluate SCOPE, we conducted experimental studies to address the following research questions.

\textit{RQ1}: How effectiveness is SCOPE?

\textit{RQ2}: How efficiency is SCOPE?

\textit{RQ3}: How does each step contribute to SCOPE?

\textit{RQ4}: How does different settings impact the result of SCOPE?

\subsection{Experiment Setup}
\subsubsection{Dataset and baselines}
The experiments are conducted on datasets of Loghub-2.0 \cite{loghub} from the LogPAI team \cite{Zhu2019}. Loghub-2.0 contains ground-truth templates of 14 log datasets which covers a wide range of operational contexts, including distributed systems, supercomputing environments, operating systems, mobile platforms, and standalone software. On average, each dataset in Loghub-2.0 contains 3.6 million log messages, all labeled with ground-truth log templates. Besides, the total number of log templates is about 3,500.


Following recent benchmark studies \cite{beck2025logparsing}, we compare our approach with a range of state-of-the-art log parsing methods. Drain \cite{drain} and Brain \cite{brain} are traditional syntax-based log parsing methods. Drain operates in an online mode, whereas Brain works offline. Both are included in our comparison as they represent the most effective approaches among non-LLM log parsers. For LLM-based log parsers, LILAC \cite{LILAC} is selected as supervised method that utilize in-context learning demonstrations for each LLM query. In contrast, LogBatcher \cite{logbatcher} and LUNAR \cite{Lunar} adopt an unsupervised strategy using a cluster-level inference method, which eliminates the need for demonstrations during LLM inference.

\subsubsection{Evaluation Metrics}
\begin{itemize}[leftmargin=*]

\item Group Accuracy (GA) \cite{Zhu2019} is a widely used metric for evaluating the effectiveness of log parsing approaches. It measures the proportion of correctly parsed logs, where a log is considered correctly parsed if it is grouped with other logs that share the same log template or pattern. GA evaluates how accurately a log parser can classify logs into distinct groups based on their structure and content, which is essential for understanding and processing system logs.


\item Parsing Accuracy (PA) \cite{Logram} is a fundamental metric for evaluating the performance of log parsing approaches. It measures the proportion of log messages that are parsed into the correct template, where a log is considered correctly parsed if the extracted template exactly matches the ground-truth template assigned to that log.

\item F1 score of Grouping Accuracy (FGA) \cite{jiang2024large} is a template-level metric that measures the ratio of correctly grouped templates. It is computed as the harmonic mean of precision and recall of grouping accuracy, where the template is considered as correct if log messages of the predicted template have the same group of log messages as the ground-truth.

\item F1 score of Template Accuracy (FTA) \cite{khan2022guidelines} is a template-level accuracy computed as the harmonic mean of precision and recall of Template Accuracy. A template is regarded as correctly identified if and only if log messages of the parsed template share the same ground-truth template and all tokens of the template are the same as those of the ground-truth template.
\end{itemize}

\subsubsection{Configurations}
The experiments were conducted on a Linux server featuring an Intel(R) Xeon(R) CPU running at 2.50 GHz on Rocky OS. We implement SCOPE based on Python 3.11.8. For comparison, we implemented various methods from public available GitHub repositories based on the default configurations. We use Qwen3-235B-A22B \cite{qwen3} as default LLM, which is open-source and has demonstrated excellent reasoning capability. 

\subsection{RQ1: How effectiveness is SCOPE?}
Table~\ref{tab:ga-pa-comparsion} presents a comparative analysis of multiple log parsing methods across 14 datasets in terms of GA, PA, FGA, and FTA. For each dataset, the highest accuracy for each metric is highlighted in bold. The experimental results demonstrate that SCOPE consistently outperforms all baselines, achieving the best overall average performance across the four metrics: GA (\textbf{96.8\%}), PA (\textbf{90.4\%}), FGA (\textbf{93.5\%}), and FTA (\textbf{83.4\%}).

Among traditional non-LLM log parsers, SCOPE exhibits substantial improvements over Drain and Brain across all evaluation metrics. Specifically, SCOPE achieves average gains of 16.1\% and 49.5\% in the group-related metrics (GA and FGA), and 94.8\% and 167.7\% in the parsing ability metrics (PA and FTA), respectively. These results highlight the superior effectiveness of LLM-based semantic parsing compared with traditional heuristic-driven methods.
When compared to LLM-based baseline parsers, SCOPE also demonstrates strong competitiveness, achieving average improvements of 3.8\%, 13.5\%, 1.5\%, and 12.0\% in GA, PA, FGA, and FTA, respectively, over LILAC, LogBatcher, and LUNAR. In particular, relative to the supervised approach LILAC, SCOPE achieves notable gains of 11.7\% and 12.6\% in PA and FTA, respectively. This indicates that SCOPE’s contrastive log-based template extraction is more effective at identifying correct templates than reliance on labeled examples, which are often unavailable in real-world systems and tend to lose generalization power when encountering new log patterns. Compared with LogBatcher and LUNAR, both of which adopt cluster-level inference strategies, SCOPE achieves average improvements of 3.2\% and 14.4\% in GA and PA, respectively. For the most stringent and comprehensive metric, FTA, SCOPE surpasses them by 11.7\%, underscoring its superior performance achieved through a carefully designed methodological framework.

\begin{table*}[t]
\centering
\caption{Accuracy comparison with the state-of-the-art log parsers. The best results are in bold.}
\label{tab:ga-pa-comparsion}
\renewcommand{\arraystretch}{1.3} 
\resizebox{\textwidth}{!}{
\begin{tabular}{c||c|cccccccccccccc|c}
\hline
\textbf{Method} & 
\textbf{Metric} & 
\textbf{HDFS} & 
\textbf{Hadoop} & 
\textbf{Spark} & 
\textbf{Zookeeper} & 
\textbf{BGL} & 
\textbf{HPC} & 
\textbf{Thunderbird} & 
\textbf{Linux} & 
\textbf{HealthApp} & 
\textbf{Apache} & 
\textbf{Proxifier} & 
\textbf{OpenSSH} & 
\textbf{OpenStack} & 
\textbf{Mac} &
\textbf{Avg.} \\ 
\hline
\multicolumn{17}{c}{\cellcolor{lightgray}\textbf{Non-LLM Log Parsers}} \\ 
\hline
\multirow{4}{*}{Drain} 
& GA & 
0.998 & 
0.948 & 
0.920 & 
0.967 & 
0.963 & 
0.887 & 
0.957 & 
0.422 & 
0.901 & 
\textbf{1.000} & 
0.765 & 
0.789 & 
0.224 & 
0.814 & 
0.825 \\
& PA & 
0.999 & 
0.613 & 
0.398 & 
0.799 & 
0.479 & 
0.662 & 
0.180 & 
0.217 & 
0.375 & 
0.978 & 
0.704 & 
0.594 & 
0.105 & 
0.392 & 
0.535 \\
& FGA & 
0.935 & 
0.785 & 
0.861 & 
0.904 & 
0.624 & 
0.309 & 
0.237 & 
0.778 & 
0.010 & 
\textbf{1.000} & 
0.206 & 
0.872 & 
0.007 & 
0.229 & 
0.554 \\
& FTA & 
0.609 &
0.384 &
0.412 &
0.614 &
0.193 &
0.152 &
0.071 &
0.259 &
0.004 &
0.517 &
0.176 &
0.487 &
0.002 &
0.069 &
0.282 \\
\hline
\multirow{4}{*}{Brain} 
& GA & 
0.960 &
0.563 &
0.972 &
0.993 &
0.940 &
0.800 &
0.792 &
0.790 &
0.979 &
0.997 &
0.521 &
0.663 &
\textbf{1.000} &
0.834 &
0.843 \\
& PA & 
0.929 &
0.143 &
0.393 &
0.822 &
0.402 &
0.663 &
0.261 &
0.010 &
0.175 &
0.287 &
0.703 &
0.481 &
0.141 &
0.325 &
0.410 \\
& FGA & 
0.759 &
0.528 &
0.208 &
0.798 &
0.756 &
0.447 &
0.748 &
0.751 &
0.872 &
0.933 &
0.737 &
0.759 &
\textbf{1.000} &
0.754 &
0.718 \\
& FTA & 
0.621 &
0.200 &
0.010 &
0.601 &
0.197 &
0.213 &
0.274 &
0.275 &
0.339 &
0.467 &
0.737 &
0.345 &
0.292 &
0.294 &
0.348 \\
\hline
\multicolumn{17}{c}{\cellcolor{lightgray}\textbf{LLM-based Supervised Log Parsers}} \\ 
\hline

\multirow{4}{*}{LILAC} 
& GA & 
\textbf{1.000} & 
0.926 & 
0.999 & 
\textbf{1.000} & 
0.911 & 
0.869 & 
0.794 & 
0.825 & 
\textbf{1.000} & 
\textbf{1.000} & 
\textbf{1.000} & 
0.748 & 
\textbf{1.000} & 
0.814 & 
0.920 \\
& PA & 
0.947 & 
0.732 & 
0.689 & 
0.685 & 
0.904 & 
0.937 & 
0.533 & 
0.814 & 
0.604 & 
0.972 & 
\textbf{1.000} & 
\textbf{0.999} & 
0.952 & 
0.564 & 
0.809 \\
& FGA & 
0.968 & 
0.941 & 
0.897 & 
\textbf{0.967} & 
0.885 & 
0.861 & 
0.859 & 
0.861 & 
0.971 & 
\textbf{1.000} & 
\textbf{1.000} & 
0.877 & 
\textbf{1.000} & 
0.871 & 
0.926 \\    
& FTA & 
0.710 & 
0.695 & 
0.636 & 
0.835 & 
0.721 & 
0.764 & 
0.557 & 
0.644 & 
0.775 & 
0.862 & 
\textbf{1.000} & 
0.849 & 
0.833 & 
0.487 & 
0.741 \\
\hline
\multicolumn{17}{c}{\cellcolor{lightgray}\textbf{LLM-based Unsupervised Log Parsers}} \\ 
\hline
\multirow{4}{*}{LogBatch} 
& GA & 
\textbf{1.000} & 
0.980 & 
0.924 & 
0.993 & 
0.990 & 
0.907 & 
\textbf{0.985} & 
0.752 & 
0.983 & 
\textbf{1.000} & 
\textbf{1.000} & 
0.624 & 
\textbf{1.000} & 
0.885 & 
0.943 \\
& PA &
\textbf{1.000} & 
0.238 & 
0.704 & 
0.655 & 
0.894 & 
0.892 & 
0.397 & 
0.655 & 
0.955 & 
0.978 & 
0.526 & 
0.772 & 
\textbf{0.989} & 
0.495 & 
0.725 \\
& FGA &
\textbf{1.000} & 
0.904 & 
0.914 & 
0.865 & 
0.945 & 
0.889 & 
\textbf{0.927} & 
0.953 & 
0.980 & 
\textbf{1.000} & 
\textbf{1.000} & 
0.717 & 
\textbf{1.000} & 
0.894 & 
0.928 \\
& FTA &
\textbf{1.000} & 
0.265 & 
0.657 & 
0.615 & 
0.766 & 
0.800 & 
0.563 & 
0.730 & 
0.725 & 
0.667 & 
0.875 & 
0.755 & 
\textbf{0.953} & 
0.540 & 
0.708 \\
\hline
\multirow{4}{*}{LUNAR} 
& GA & 
\textbf{1.000} & 
0.941 & 
0.975 & 
0.993 & 
0.955 & 
0.864 & 
0.865 & 
0.830 & 
\textbf{1.000} & 
\textbf{1.000} & 
0.989 & 
0.780 & 
\textbf{1.000} & 
0.877 & 
0.934 \\
& PA &
\textbf{1.000} & 
0.837 & 
0.995 & 
0.851 & 
\textbf{0.984} & 
0.990 & 
0.595 & 
0.735 & 
\textbf{0.962} & 
0.999 & 
\textbf{1.000} & 
0.722 & 
0.904 & 
0.583 & 
0.868 \\
& FGA &
0.968 & 
0.926 & 
0.888 & 
0.885 & 
0.873 & 
0.833 & 
0.869 & 
0.874 & 
0.971 & 
\textbf{1.000} & 
0.870 & 
0.923 & 
\textbf{1.000} & 
0.863 & 
0.910 \\
& FTA &
0.946 & 
0.702 & 
0.657 & 
0.812 & 
\textbf{0.789} & 
\textbf{0.833} & 
\textbf{0.574} & 
0.721 & 
0.843 & 
0.897 & 
0.957 & 
\textbf{0.923} & 
0.875 & 
0.526 & 
0.790 \\
\hline
\multicolumn{17}{c}{\cellcolor{lightgray}\textbf{Our proposed method}} \\ 
\hline
\multirow{4}{*}{\textbf{SCOPE}}
& GA & 
\textbf{1.000} & 
\textbf{0.988} & 
\textbf{1.000} & 
0.991 & 
0.962 & 
\textbf{0.972} & 
0.908 & 
\textbf{0.925} & 
0.937  & 
0.993  & 
\textbf{1.000}  & 
\textbf{0.928}  & 
\textbf{1.000} & 
\textbf{0.948} & 
\textbf{0.968} \\
& PA &
\textbf{1.000}  & 
\textbf{0.894} & 
\textbf{0.997} & 
0.798 & 
0.949 & 
\textbf{0.990} & 
\textbf{0.602} & 
\textbf{0.884} & 
0.953 & 
0.991 & 
\textbf{1.000} & 
0.839 & 
0.975 & 
\textbf{0.780} & 
\textbf{0.904} \\
& FGA &
0.968 & 
\textbf{0.948} & 
\textbf{1.000} & 
0.918 & 
\textbf{0.962} & 
0.851 & 
0.877 & 
\textbf{0.974} & 
0.928 & 
0.844 & 
\textbf{1.000} & 
0.900 & 
\textbf{1.000} & 
\textbf{0.926} & 
\textbf{0.935} \\
& FTA &
0.860 & 
\textbf{0.733} & 
\textbf{0.861} & 
0.787 & 
\textbf{0.862} & 
\textbf{0.835} & 
\textbf{0.685} & 
\textbf{0.828} & 
\textbf{0.926} & 
0.854 & 
\textbf{1.000} & 
0.825 & 
0.833 & 
\textbf{0.784} & 
\textbf{0.834} \\
\hline
\end{tabular}}
\end{table*}

\noindent
\setlength{\fboxsep}{4pt} 
\colorlet{mygray}{gray!15} 

\begin{center}
\fcolorbox{black}{mygray}{%
  \parbox{0.95\columnwidth}{%
    \small 
    \textbf{Answer to RQ1:} SCOPE outperforms all baseline log parsers across all evaluation metrics, 
    regardless of whether they are non-LLM or LLM-based methods. Moreover, as an online approach, 
    SCOPE even surpasses offline methods that leverage LLMs. These results clearly demonstrate 
    that SCOPE’s novel design advances the state of the art in log parsing effectiveness.
  }
}
\end{center}

\subsection{RQ2: How efficiency is SCOPE?}
\subsubsection{Efficiency Analysis}
\begin{figure}[htbp]
\centerline{\includegraphics[width=\linewidth]{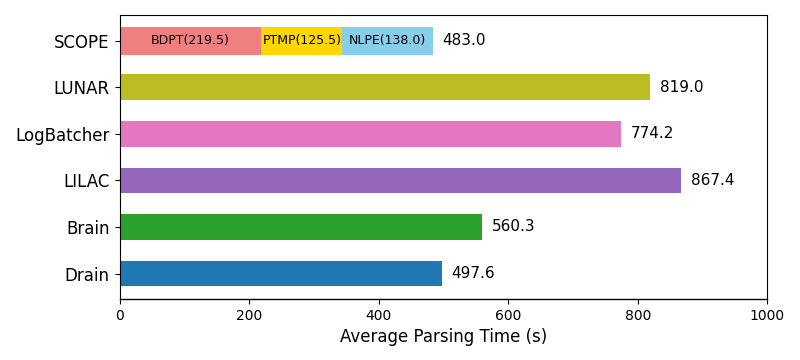}}
\caption{Efficiency comparison with baseline parsers. 
SCOPE achieves the highest efficiency, with BDPT (219.5\,s), PTMP (125.5\,s), and NLPE (138\,s).}
\label{fig:timebar}
\end{figure}

Efficiency in log parsing is critical for software-intensive systems, as it directly affects the speed of failure detection and recovery. In this section, we evaluate the time efficiency of SCOPE in comparison with baseline log parsers by applying them to large-scale datasets from Loghub-2.0 under default parameter settings. Specifically, we measure and compare the parsing time of SCOPE and the baseline parsers across all datasets. From the analysis of individual datasets, the parsing time of each method is roughly proportional to the number of log templates in the dataset, while the average parsing time across all datasets is summarized in Fig.~\ref{fig:timebar}.

From the figure, we observe that SCOPE exhibits the lowest time consumption among all log parsing methods. In detail, SCOPE requires 483 seconds to parse an average of 3.6 million log messages, outperforming Drain, the most efficient syntax-based log parser currently available, which takes 497.6 seconds. Considering only the processing time for parsing cache, SCOPE spends a total of 345 seconds on the BDPT and PTMP modules, reducing the time cost by approximately 30\% compared with Drain. The superior efficiency of SCOPE primarily stems from the bi-directional tree design in BDPT, which improves the hit rate of log–template matching in the early phase. Moreover, the dynamic tree depth design of BDPT significantly reduces the number of candidate template matches at the leaf nodes compared with the fixed-depth design of Drain, thereby improving the overall matching efficiency.

When compared with LLM-based log parsers, including LILAC, LogBatcher, and LUNAR, SCOPE achieves approximately 41\% higher efficiency. Our analysis indicates these methods spend a significantly larger portion of their execution time on LLM invocations, This is largely because they rely on querying the LLM to obtain new templates, either by extracting from pre-clustered groups, as in LogBatcher and LUNAR, or via in-context learning in LILAC, resulting in the number of API calls being equal to or even exceeding the number of templates when parsing errors occur, thereby further increasing the required LLM invocations. In contrast, SCOPE invokes the LLM only when the preceding modules fail to extract a template, thereby minimizing the overhead of LLM invocations. Specifically, as shown in Table~\ref{tab:llm-usage}, SCOPE requires only about 15\% as many LLM calls, demonstrating its highly efficient utilization of LLM resources.

\subsubsection{Cost Analysis}
LLM-based approaches incur additional costs due to their reliance on LLM API invocations, which are provided by commercial service providers such as OpenAI and billed based on token usage. As a result, each LLM query introduces a financial cost proportional to the number of input and output tokens. In this section, we conduct a financial cost analysis of SCOPE in comparison with other LLM-based baseline methods to assess its practical feasibility. Specifically, we report (1) the average accuracy across four evaluation metrics, (2) the total number of tokens consumed per dataset, (3) the number of LLM invocations per dataset, and (4) the total dollar cost per dataset, as summarized in Table~\ref{tab:llm-usage}. SCOPE consumes an average of 98.4K tokens per dataset, which is lower than LUNAR (142.8K) but slightly higher than LILAC (82.9K) and LogBatcher (65.8K). This is mainly attributed to the Chain-of-Thought (CoT) prompting strategy adopted in SCOPE, which increases token usage. However, SCOPE eliminates the costs of human annotation and demonstration selection required by LILAC, making it a label-free alternative with substantially reduced manual effort. Compared with LogBatcher, SCOPE incurs an additional cost of approximately \$0.025 per dataset while achieving an average accuracy improvement of 8\%. This trade-off is clearly worthwhile, as the performance gains significantly outweigh the marginal cost increase. Moreover, with the continued decline in token prices, this minor disadvantage is expected to diminish over time.


\begin{table}[htbp]
\centering
\caption{Statistics of LLM invocations and tokens cost}
\begin{tabular}{lcccc}
\toprule
\textbf{Method} & \textbf{Acc.} & \textbf{\#Invocations} & \textbf{\#Tokens} & \textbf{Cost} \\
\midrule
LILAC        & 0.85 & 279.7 & 82.9K & \$0.052 \\
LogBatcher   & 0.83 & 258.2 & 65.8K & \$0.038\\
LUNAR        & 0.88 & 268.5 & 142.8K & \$0.084 \\
\cline{1-5}
SCOPE        & 0.91 & 41.7 & 98.4K & \$0.062 \\
\bottomrule
\end{tabular}
\label{tab:llm-usage}
\end{table}

\noindent
\setlength{\fboxsep}{4pt} 
\colorlet{mygray}{gray!15} 

\begin{center}
\fcolorbox{black}{mygray}{%
  \parbox{0.95\columnwidth}{%
    \small 
    \textbf{Answer to RQ2:} SCOPE demonstrates clear efficiency advantages over both non-LLM and LLM-based log parsers on large-scale datasets. It achieves the best accuracy metrics while reducing LLM invocations by a factor of six and maintaining competitive token usage. Overall, SCOPE represents the most promising approach for real-world log parsing applications.
  }
}
\end{center}

\subsection{RQ3: How does each step contribute to SCOPE?}
\subsubsection{Overall Statistics} In the end-to-end experiments of SCOPE, we collected statistics from BDPT and PTMP during the log matching and NLPE invocation processes to evaluate their respective contributions. As shown in Table~\ref{tab:component-stats}, in the log matching phase, BDPT and PTMP account for 95\% and 5\% of the operations, respectively. This is because BDPT is applied first for matching due to the high efficiency of tree traversal, allowing the majority of log messages to be captured in this phase. Moreover, its relative proportion continues to increase as the number of log messages grows. Regarding NLPE invocations, BDPT accounts for 64\% of the calls, indicating that the majority of template extraction is performed within BDPT, which demonstrates its crucial role in the overall parsing process. In terms of template extraction counts, BDPT maintains a consistent trend with NLPE invocations, accounting for 69\% overall. Among these, the forward branch and reverse branch contribute 46\% and 23\%, respectively, highlighting the important contribution of the reverse tree in capturing templates.

\begin{table}[h]
\centering
\caption{SCOPE Log parsing statistics by component}
\begin{tabular}{lccc}
\hline
\textbf{Comp.} & \textbf{\#Log Match} & \textbf{\#NLPE Inv.} & \textbf{\#Templates Extr.} \\
\hline
BDPT & 95\% & 64\% & \makecell[l]{Fwd: 46\% \\ Rev: 23\%} \\
PTMP & 5\% & 36\% & 31\% \\
\hline
\end{tabular}
\label{tab:component-stats}
\end{table}

We then conducted an ablation study on the individual components. As shown in Table~\ref{tab:ablation}, the results clearly indicate that the removal of any component leads to a measurable decrease in performance.

\begin{table}[htbp]
\centering
\caption{SCOPE Perfermance under different settings}
\begin{tabular}{lccc}
\toprule
\textbf{} & \textbf{GA} & \textbf{PA} & \textbf{LLM calls} \\
\midrule
Full SCOPE        & 
0.968 & 
0.904 & 
41.7 
\\
{w/o} {NLPE}      & 
0.834~($\downarrow$15.8\%) & 
0.780~($\downarrow$15.0\%) & 
0 
\\
{ -w/o} {LLM} & 
0.901~($\downarrow$6.7\%) & 
0.833~($\downarrow$7.1\%) & 
0 
\\
{ -w/o} {POS} & 
0.968~($\downarrow$0.0\%) & 
0.903~($\downarrow$0.1\%) & 
63.8~($\uparrow$53.1\%) 
\\
{w/o} {PTMP}      & 
0.966~($\downarrow$2.5\%) & 
0.903~($\downarrow$1.6\%) &
35.0~($\downarrow$16.3\%) 
\\
{w/o} {BDPT}      & 
0.967~($\downarrow$0.1\%) & 
0.890~($\downarrow$1.4\%) & 
40.7~($\downarrow$3.1\%) 
\\
\bottomrule
\end{tabular}
\label{tab:ablation}
\end{table}


\subsubsection{{w/o} {NLPE}}
As shown in Table~\ref{tab:ablation}, removing NLPE results in a significant drop in performance: 15.8\% in GA and 15.0\% in PA. This decline is substantially larger than the impact of removing any other component, highlighting the critical role that NLPE plays in the log parsing process.
To further understand the contribution of the two stages within NLPE, we conduct more fine-grained ablations:
(a) {{w/o} {LLM}}. Removing only the LLM component results in a 6.7\% decrease in GA and 7.1\% in PA, demonstrating that LLM plays a significant role in enhancing parsing accuracy.
(b) {{w/o} {POS}}. Removing only the POS component, however, causes minimal change in GA and PA. This is because the LLM module compensates by handling cases originally parsed by POS. Nevertheless, removing POS leads to a 53.1\% increase in LLM API calls, indicating that POS effectively offloads a substantial number of common log parsing scenarios. 
\subsubsection{{w/o} {PTMP} }
PTMP is activated when BDPT template matching fails, particularly in cases where variables appear in both halves of a log message. Removing PTMP leads to a 2.5\% drop in GA and a 1.6\% drop in PA, as some templates can no longer be correctly identified. Meanwhile, LLM call volume decreases by 16.3\% due to the absence of PTMP-triggered NLPE invocations. These results indicate that PTMP effectively addresses the limitations of tree-based log parsing and serves as an indispensable component in enhancing SCOPE’s template extraction accuracy.
\subsubsection{{w/o} {BDPT}}
The ablation results suggest that removing BDPT does not notably affect SCOPE’s GA and PA, as PTMP fully handles template matching in its absence. However, this should not be interpreted as BDPT being unnecessary.
In fact, BDPT performs efficient tree-based traversal and matching in the majority of cases, as evidenced by the log match number statistics presented in \ref{tab:component-stats}. Furthermore, compared to the single-directional parse tree used in the Drain \cite{drain} log parser, our bi-directional tree captures 69\% of templates, with 23\% matched exclusively by the reverse tree.

\noindent
\setlength{\fboxsep}{4pt} 
\colorlet{mygray}{gray!15} 

\begin{center}
\fcolorbox{black}{mygray}{%
  \parbox{0.95\columnwidth}{%
    \small 
    \textbf{Answer to RQ3:} NLPE elevates SCOPE’s effectiveness to a new high. By integrating POS-based parsing with LLM-based fallback, it achieves a well-balanced trade-off between accuracy and efficiency. PTMP enhances the completeness of template matching and BDPT ensures efficient and scalable template extraction, laying the foundation for SCOPE’s practical deployment.
  }
}
\end{center}

\subsection{RQ4: How does different settings impact the result of SCOPE?}
\subsubsection{LLM model selection}
We evaluate the impact of different LLM backends by testing both open-source Qwen3 series models (including 235B, 32B, and 8B variants)\cite{qwen3} and the closed-source GPT-4o(\textit{gpt-4o-0806})\cite{gpt-4o} and Claude-3.5(\textit{claude-3-5-sonnet-20240620})\cite{anthropic2024claude3}. Detailed results are shown in Table~\ref{tab:configSetting}.


\begin{table}[htbp]
\centering
\caption{Configuration study with different LLM models}
\begin{tabular}{lccccc}
\toprule
\textbf{Model} & 
\textbf{Parameters} & 
\textbf{GA} & 
\textbf{PA} & 
\textbf{FGA} & 
\textbf{FTA} \\
\midrule
\multirow{3}{*}{Qwen3}
 & 235B &  0.968  & 0.904	& 0.935	& 0.834 \\
 & 32B  &  0.966  & 0.902  & 0.928 & 0.828 \\
 & 8B   &  0.952  & 0.882  & 0.915 & 0.814 \\
\hline
GPT-4o & \string~200B  &  0.968 & 0.902 & 0.935 & 0.830 \\
\hline
Claude-3.5 & \string~175B &  0.965 & 0.896 & 0.930 & 0.826 \\
\bottomrule
\end{tabular}
\label{tab:configSetting}
\end{table}

The results reveal that SCOPE exhibits strong robustness across LLM model scales. Even the Qwen3-8B variant achieves competitive accuracy, while GPT-4o and Claude-3.5 consistently maintains superior performance. This demonstrates SCOPE’s generalization ability and insensitivity to the choice of LLM, making it adaptable across a wide range of deployment scenarios. 

\subsubsection{LLM prompt strategy}
To evaluate the impact of the prompting strategy, we compare the performance of SCOPE using standard prompts and chain-of-thought (CoT) prompts. As shown in Table~\ref{tab:promptSetting}, CoT prompting significantly improves the average accuracy across all four metrics. Specifically, for the Qwen3 series models, the average accuracy increases by 8.2\%, with particularly notable improvements in smaller LLMs. For instance, the accuracy of the Qwen3-8B model increases by 11.2\% when CoT prompting is applied. These results highlight the importance of effective prompting strategies, especially when working with resource-constrained models.

\begin{table}[htbp]
\centering
\caption{Accuracy under different prompt strategies}
\begin{tabular}{lccc}
\toprule
\textbf{Model} & \textbf{Parameters} & \textbf{w/ CoT} & \textbf{w/o CoT} \\
\midrule
\multirow{3}{*}{Qwen3} & 235B & 0.91 & 0.86\\
& 32B & 0.90  & 0.82\\
& 8B & 0.89  & 0.78\\
\bottomrule
\end{tabular}
\label{tab:promptSetting}
\end{table}

\noindent
\setlength{\fboxsep}{4pt} 
\colorlet{mygray}{gray!15} 
\begin{center}
\fcolorbox{black}{mygray}{%
  \parbox{0.95\columnwidth}{%
    \small 
    \textbf{Answer to RQ4:} SCOPE exhibits strong robustness across different LLM model choices. With the use of CoT prompting strategy, it achieves leading performance even when using lightweight models such as Qwen3-8B. This demonstrates SCOPE's adaptability and practical deployment potential under resource-constrained settings.
  }
}
\end{center}

\section{Threats to Validity}
\textbf{External Threat.}
A key threat to external validity is the limited diversity of the evaluation datasets. The experiments are conducted on public log datasets that are relatively clean and structurally simple, which may not fully capture the complexity and heterogeneity of real-world production environments. This limitation may negatively affect the generalizability of the results. Recently, LogBase\cite{logbase} released a diverse and large-scale log dataset comprising 130 popular open-source projects and 85,300 semantically annotated log templates. Future work should evaluate SCOPE on such more diverse and large-scale log datasets to more thoroughly assess its scalability and robustness across a wide range of industrial scenarios.

\textbf{Internal Threat.}
Since LLMs may optimize their internal reasoning and skip intermediate steps even if temperature is set to zero, they can produce inconsistent outputs for the same input. This behavior introduces uncertainty that makes it difficult to isolate the performance impact of our framework design. To address this, we explicitly prompt the LLM to return step-by-step results, encouraging more consistent reasoning paths. In addition, we repeat each experiment 5 times and report the average to ensure a more reliable evaluation.


\section{Related work}

Syntax-based log parsers \cite{Logram,drain,jiang2008abstracting,brain} assume that log templates share recurring patterns that consistently appear throughout the entire log dataset. Log parsers \cite{Logram, nagappan2010abstracting,vaarandi2015logcluster} extract these templates by detecting the stable components of log messages through frequent pattern mining. For example, \textit{Brain} \cite{brain} is a log parsing method that first creates initial groups by identifying the longest common pattern in log messages, leveraging common log generation characteristics. It then refines these groups by hierarchically adding constant words using a bidirectional tree to hierarchically add constant words to the longest common pattern.
Heuristics-based log parsers \cite{drain,jiang2008abstracting} utilize distinctive features of log messages to effectively identify shared templates.
Drain \cite{drain} build hierarchical structures to filter and classify logs efficiently, avoiding full comparisons by narrowing the search space based on log characteristics (e.g., length, token position).

Semantic-based log parsers exploit the semantic distinctions between keywords and parameters, framing log parsing as a token-level classification problem. VALB \cite{variableAwareLog} utilizes a BiLSTM-CNN-CRF structure to identify the semantic category of each log token. SemParser \cite{huo2023semparser} introduces an end-to-end semantic miner and joint parser, leveraging domain knowledge to identify semantics in logs. LogPPT \cite{jiang2008abstracting} proposes an innovative approach to log parsing by utilizing template-free prompt tuning to adapt the pre-trained language model RoBERTa \cite{liu2019roberta,devlin2019bert}. While effective, current semantic-based parsers often incur significant training costs, such as building models from the ground up or fine-tuning pre-trained models using annotated data which is both limited in availability and expensive to produce \cite{le2023log}.

LLM-based log parsers have recently emerged to address the limitations of traditional approaches. By leveraging the rich pre-trained knowledge and powerful in-context learning (ICL) abilities of large language models (LLMs) \cite{xia2024fuzz4all}, researchers have achieved promising log parsing performance \cite{guo2023images,lewis2019bart,zhou2024leveraging,zhi2024llm}. For example, DivLog \cite{divlog} and LILAC \cite{LILAC} adopt an ICL method, providing labeled demonstrations in the LLM prompt to improve accuracy. While such supervised methods deliver strong performance, their practicality is limited because labeled examples are often difficult to obtain in real-world systems. Recently, the Cluster-Level Inference method has gained popularity. This approach first clusters logs and then extracts templates at the cluster level. As an unsupervised LLM-based parsing paradigm, it eliminates the need for labeled examples and achieves stronger generalization to diverse. For example, LogBatcher \cite{logbatcher} and LUNAR \cite{Lunar} both achieve excellent results under this paradigm, while also lowering the cost and constraints, making them more suitable for deployment in real-world environments.

\section{Conclusion}
In this work, we propose SCOPE, an online unsupervised log parser that integrates syntactic-semantic collaboration with a bidirectional tree for self-correcting log parsing.
SCOPE integrates three key components: 1) BDPT is a bidirectional parse tree that enhances token matching by effectively handling variations in both the front and back parts of log entries. 2) PTMP is a priority-based template match pool that serves as a global fallback matcher when BDPT fails. It ensures robust coverage of complex and uncommon log patterns. 3) NLPE is an NLP-based extractor that utilizes syntactic and semantic features to precisely distinguish constants from variables, thereby enhancing template extraction.
Extensive experiments on widely-used benchmark datasets demonstrate that SCOPE significantly outperforms state-of-the-art parsers in parsing accuracy while maintaining competitive processing efficiency.

\section*{Data Availability}
The code and data are available: \url{https://github.com/Fan-Dongyi/scopeLog}


\bibliographystyle{ACM-Reference-Format}
\bibliography{ICPC2026-SCOPE}

\end{document}